\documentclass[preprint,12pt,times]{elsarticle}
\usepackage{color,soul}
\usepackage{dsfont,mathrsfs}
\usepackage{amsmath,amssymb,amsfonts}
\usepackage{amsthm}
\usepackage{algorithm}
\usepackage{booktabs} 
\usepackage{graphicx}
\usepackage{subcaption}
\usepackage{hyperref}
\usepackage{bm}

\newproof{pf}{Proof}

\journal{Aerospace Science and Technology (AESCTE)}

\begin{document}

\begin{frontmatter}

\title{Imitation learning-based spacecraft rendezvous and docking method with Expert Demonstration}

\author[1]{Shibo Shao}
\ead{22b904044@stu.hit.edu.cn}

\author[1,2]{Dong Zhou\corref{Corauthor}}
\ead{dongzhou@hit.edu.cn}

\author[1]{Guanghui Sun}
\ead{guanghuisun@hit.edu.cn}

\author[1]{Liwen Zhang}
\ead{liwenzhang@stu.hit.edu.cn}

\author[1]{Minxuan Jiang}
\ead{2022112926@stu.hit.edu.cn}
%

%
%
\address[1]{Department of Control Science and Engineering, Harbin Institute of Technology, Harbin 150001, China}
\address[2]{Department of Mechanical and Automation Engineering, The Chinese University of Hong Kong, Hong Kong 999077, China}
\cortext[Corauthor]{Dong Zhou}

\begin{abstract}
Existing spacecraft rendezvous and docking control methods largely rely on predefined dynamic models and often exhibit limited robustness in realistic on-orbit environments. To address this issue, this paper proposes an Imitation Learning-based spacecraft rendezvous and docking control framework (IL-SRD) that directly learns control policies from expert demonstrations, thereby reducing dependence on accurate modeling. We propose an anchored decoder target mechanism, which conditions the decoder queries on state-related anchors to explicitly constrain the control generation process. This mechanism enforces physically consistent control evolution and effectively suppresses implausible action deviations in sequential prediction, enabling reliable six-degree-of-freedom (6-DOF) rendezvous and docking control. To further enhance stability, a temporal aggregation mechanism is incorporated to mitigate error accumulation caused by the sequential prediction nature of Transformer-based models, where small inaccuracies at each time step can propagate and amplify over long horizons. Extensive simulation results demonstrate that the proposed IL-SRD framework achieves accurate and energy-efficient model-free rendezvous and docking control. Robustness evaluations further confirm its capability to maintain competitive performance under significant unknown disturbances. The source code is available at https://github.com/Dongzhou-1996/IL-SRD. 

\end{abstract}

\end{frontmatter}

\section{Introduction} \label{intro}
Spacecraft rendezvous and docking refers to the process in which a deputy spacecraft autonomously regulates its translational and rotational states to safely dock with a chief spacecraft ~\cite{xie2021guidance, sharma2024advancing}. This capability is fundamental to a wide range of on-orbit missions, including space station assembly and maintenance, on-orbit servicing, and space debris removal.

Early spacecraft rendezvous and docking missions relied entirely on manual control \cite{grimwood1969project, young1970apollo}, which was shown to be hazardous, cognitively demanding, and unreliable \cite{cok1989omv, zhou2012new, mokuno1995development, zhou2014design}. To alleviate these limitations, teleoperation-based control was subsequently developed and adopted by several space agencies \cite{cok1989omv, zhou2012new, mokuno1995development, zhou2014design}. However, teleoperation-based approaches are fundamentally constrained by communication latency, which degrades control precision and responsiveness. Moreover, high operational costs, unstable communication links, and incomplete or asymmetric information further hinder the scalability and practical deployment of teleoperation-based spacecraft rendezvous and docking systems.

To further improve control efficiency and mission success rates, optimization-based automatic control methods were developed, with human operators retained as backups for contingency scenarios. Classical controllers, such as PID, have been applied to spacecraft rendezvous and docking tasks \cite{KOSARI2017293, kosari2017optimal1}, while Model Predictive Control (MPC) has been extensively studied for its ability to explicitly handle safety and operational constraints \cite{weiss2015model, di2012model, li2017model}. In addition, Sliding Mode Control (SMC) has attracted considerable attention due to its robustness against model uncertainties and external disturbances \cite{zhang2016disturbance, 10055300, li2018sliding}. Nevertheless, these methods generally rely on accurate dynamic models. In the absence of careful modeling and parameter tuning, performance degradation may occur, leading to reduced control accuracy, increased fuel consumption, or even mission failure.

With the advancement of deep learning, Deep Neural Networks (DNNs) and Deep Reinforcement Learning (DRL) have been increasingly applied to spacecraft rendezvous and docking control, either to enhance traditional control frameworks \cite{ramezani2025mpc, zhao2025fault} or to directly learn autonomous control policies through interaction with the environment \cite{zhou2023deep, hovell2021deep, zhou2022space, yuan2022deep, di2025trajectory}. In addition to the requirement for accurate system modeling, DRL-based control methods are also highly sensitive to reward function design. Improperly designed reward functions may lead to misaligned or unsafe policies, resulting in suboptimal performance, excessive fuel consumption, or even instability and collision risks during testing.

\begin{figure}[ht]
    \centering
    \includegraphics[width=\linewidth, trim= 20 20 20 10]{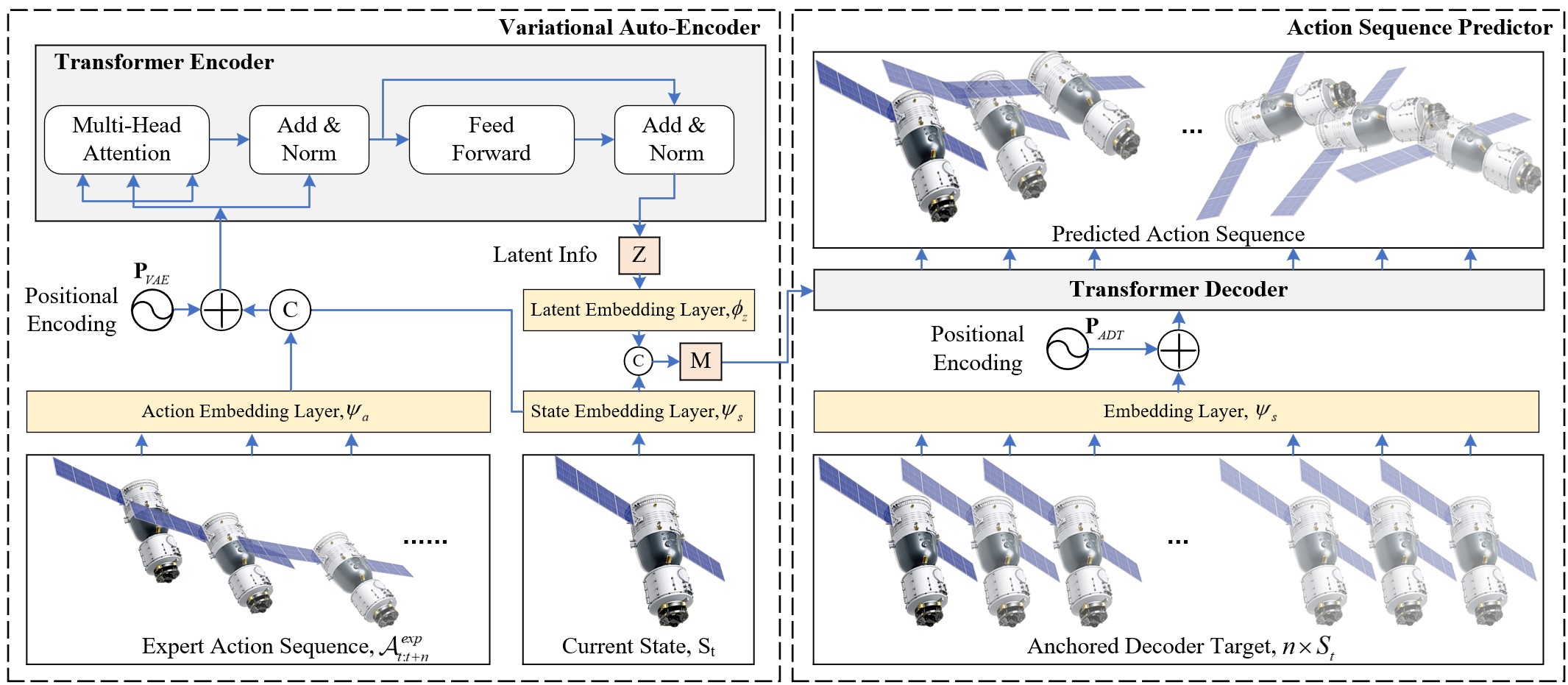}
    \caption{Network structure of IL-SRD}
    \label{fig:network}
\end{figure}

To address these challenges, we propose an Imitation Learning-based Spacecraft Rendezvous and Docking (IL-SRD) control algorithm, as illustrated in Fig. \ref{fig:network}. Unlike conventional spacecraft rendezvous and docking control methods, the proposed approach directly learns a control policy from expert demonstrations, thereby alleviating the reliance on accurate spacecraft dynamics. However, spacecraft rendezvous and docking is inherently a long-horizon and 
strongly coupled control problem, for which naive imitation learning methods such as Behavioral Cloning (BC) \cite{bain1995framework} are inadequate, as they fail to capture the temporal dependencies between states and action sequences. To address this limitation, a Transformer-based architecture \cite{vaswani2017attention} is employed to model long-horizon temporal dependencies in control sequence generation.

To address the physical and safety-critical requirements of spacecraft rendezvous and docking control, we introduce an anchored decoder target (ADT) mechanism that explicitly constrains predicted action sequences using state-related anchors. This design enforces physical consistency and actuator feasibility during control generation, effectively suppressing irrational or unsafe action deviations. Inspired by the ACT algorithm \cite{zhao2023learning}, we further incorporate a temporal aggregation mechanism to enhance long-term control stability and mitigate error accumulation in sequential prediction. Together, these designs ensure smooth and stable control execution throughout the rendezvous and docking process.

The main contributions are summarized as follows:
\begin{itemize} 
\item We propose an imitation learning-based spacecraft rendezvous and docking control algorithm, IL-SRD, that explicitly captures long-horizon temporal dependencies in expert demonstrations, enabling the direct learning of a near-optimal 6-DOF control policy.
\item We propose an ADT mechanism to resolve physical inconsistency in predicted action sequences and introduce a temporal aggregation mechanism to resolve error accumulation in sequential prediction, thereby enabling more accurate and stable long-horizon control.
\item Extensive experimental evaluations demonstrate that the proposed method significantly outperforms classical controllers, DRL-based approaches, and vanilla BC. Ablation studies further validate the effectiveness of each key component.
\end{itemize}
 

\section{Related Works} \label{sec: rw}
Classical model-based control methods for spacecraft rendezvous and docking have been extensively studied. Kosari \textit{et al.} utilize fuzzy PID controllers combined with genetic algorithms to handle the rotational \cite{KOSARI2017293} and translational \cite{kosari2017optimal1} phases of rendezvous and docking, respectively. However, due to the inherent limitations of PID control in handling strong system coupling and explicit constraints, such approaches are generally less suitable for complex rendezvous and docking missions. 

Model Predictive Control (MPC) is widely recognized for its capability to explicitly enforce multiple constraints while preserving control performance. Recent work proposes a robust control framework that integrates trajectory planning, robust tracking, and constraint handling within a unified MPC-based architecture \cite{kang2025controller}. By incorporating an adaptive smooth controller and a controller-matching strategy, the robustness of the MPC is enhanced when constraints are inactive, while improved computational efficiency is maintained. Iskender \textit{et al.} \cite{iskender2019dual} adopt dual quaternions to provide a unified representation of translational and rotational dynamics. As a result, the coupled position–attitude dynamics can be optimized simultaneously while explicitly handling actuator and safety constraints.

On the other hand, Capello \textit{et al.} \cite{capello2017sliding} employ sliding mode control for spacecraft rendezvous and docking due to its inherent robustness to modeling uncertainties and external disturbances. By constructing nonlinear sliding surfaces based on relative translational and rotational dynamics, the proposed approach guarantees finite-time convergence and maintains strong robustness throughout the rendezvous maneuver.

All the methods mentioned above fundamentally rely on accurate dynamic modeling of the system. PID controllers are typically designed based on simplified or decoupled models, which limits their applicability to highly coupled rendezvous and docking dynamics. In contrast, MPC and SMC explicitly incorporate system dynamics into controller design and optimization, rendering their performance highly sensitive to modeling fidelity. Consequently, substantial effort is required to account for time-varying and unknown disturbances in order to ensure reliable control performance. In contrast, the method proposed in this paper does not require explicit knowledge of the system dynamics; instead, long-horizon temporal dependencies are directly learned from expert demonstrations.

DRL-based control algorithms for spacecraft rendezvous and docking significantly differ from classical model-based approaches, in which control policies are learned through direct interaction with the environment via a trial-and-error process. The performance of such policies is therefore mainly governed by the design of the reward function. Yang \textit{et al.} \cite{yang2025spacecraft} evaluate multiple DRL-based algorithms for autonomous spacecraft rendezvous using a nonlinear dynamic model formulated on the Lie Special Euclidean Group in 3D. The reward function combines dense step-wise feedback, an integral penalty to suppress accumulated tracking errors over a finite horizon, and a terminal reward to promote successful docking. Qu \textit{et al.} \cite{qu2022spacecraft} focus on translational rendezvous with collision avoidance, where the reward function integrates multiple objectives, such as time efficiency, safety constraints, state regulation, and fuel consumption. These studies indicate that DRL-based spacecraft rendezvous and docking control methods suffer from several inherent limitations, including a strong dependence on reward functions, high sensitivity to hyperparameter selection, and long training durations before convergence. In contrast, the training period of IL-SRD is significantly reduced with a simple and well-defined loss function. 

In contrast, imitation learning (IL) typically does not require explicit system modeling or prolonged training. The control policy is directly learned from expert demonstrations. Federici \textit{et al.} \cite{federici2021deep} combine deep reinforcement learning and behavioral cloning to achieve translational position and velocity control under visibility cone constraints. Although only translational dynamics are considered, their results demonstrate the feasibility of IL-based approaches for spacecraft rendezvous and docking tasks. Kim \textit{et al.} \cite{kim2025constrained} employ imitation learning to train a Time Shift Governor, a parameter-regulating mechanism that enforces state and control constraints while simultaneously reducing the onboard computational burden. More recently, Posadas-Nava \textit{et al.} \cite{posadas2025action} propose an imitation learning framework that leverages both numerical state information and visual observations to replicate expert guidance and control behaviors. 

Compared with the methods discussed above, the proposed approach learns the control policy directly from expert trajectories, without relying on meticulous system dynamics modeling or handcrafted reward function design. By incorporating the proposed ADT and temporal aggregation mechanism, the learned policy remains physically consistent and avoids error accumulation in long-horizon prediction. As a result, IL-SRD achieves accurate six-degree-of-freedom spacecraft rendezvous and docking control in a fully model-free manner. Moreover, the proposed framework effectively captures the temporal dependencies embedded in expert demonstrations and is able to reproduce comparable performance under both nominal and disturbed conditions, highlighting its robustness and practical applicability to realistic on-orbit scenarios.

\section{Methodology}
\subsection{Problem Statement}
In this section, the spacecraft rendezvous and docking control will be explained in detail with the dynamic models we established for collecting expert demonstrations. 
\begin{figure}[htb]
    \centering
    \includegraphics[width=\linewidth, trim=30 50 60 150]{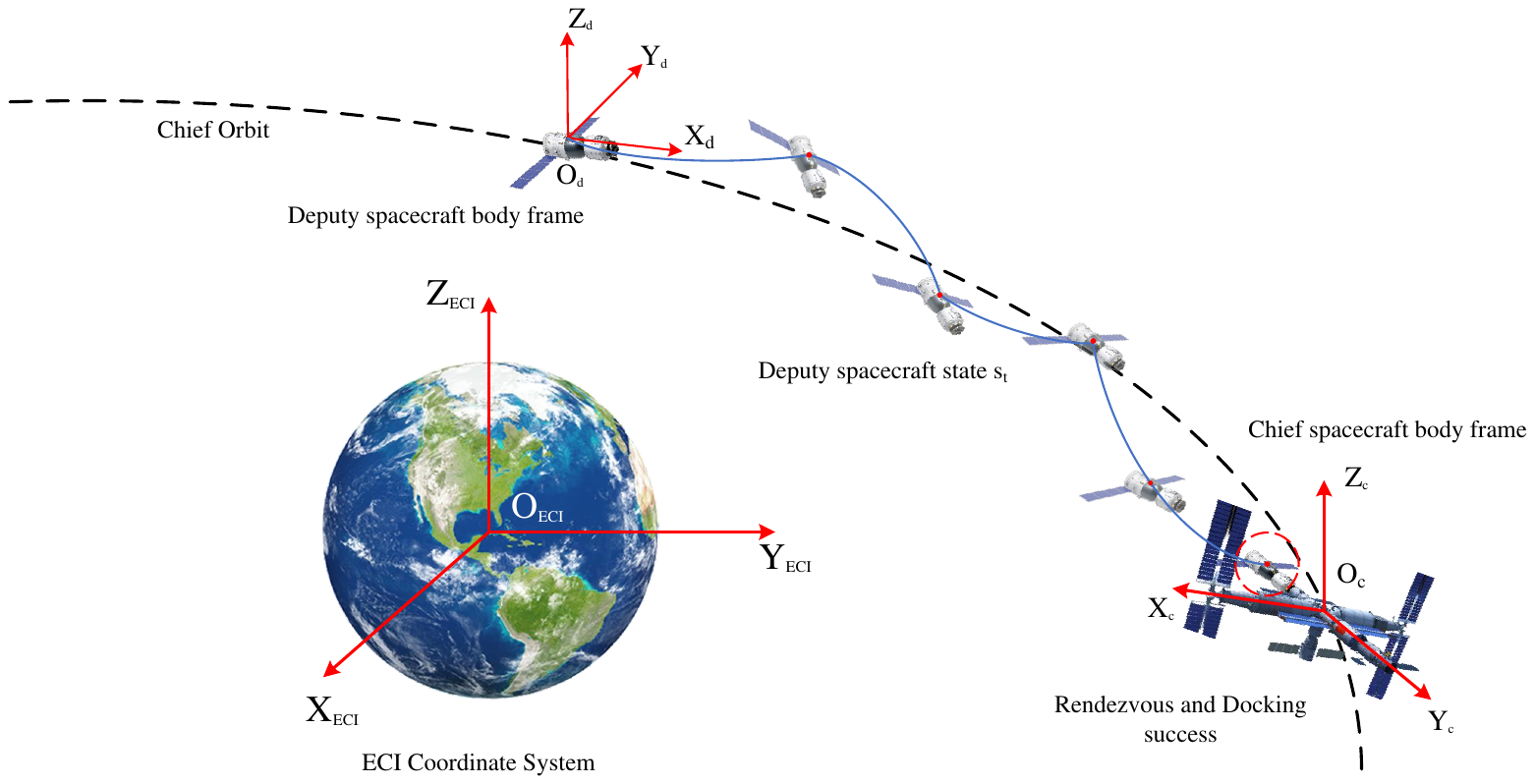}
    \caption{Overview of the on-orbit spacecraft rendezvous and docking control}
    \label{fig:dynamic}
\end{figure}
Fig. \ref{fig:dynamic} illustrates the spacecraft rendezvous and docking maneuver process. Three coordinate systems are employed in this paper: the Earth-Centered Inertial (ECI) frame $\mathcal{F}_{ECI}$, the chief spacecraft body frame $\mathcal{F}_{c}$, and the deputy spacecraft body frame $\mathcal{F}_{d}$. The chief spacecraft is assumed to undergo free-floating rigid-body motion. Therefore, the body frame $\mathcal{F}_{c}$ is coincident with the Local Vertical Local Horizontal (LVLH) frame. In the following equations, the LVLH frame is used interchangeably to represent $\mathcal{F}_{c}$.

The primary purpose of the ECI coordinate system is to derive the initial positions and velocities of the chief and deputy spacecraft from their Classical Orbital Elements (COEs). To simplify the dynamic model, a Keplerian circular orbit is adopted instead of an elliptical orbit. The attitude of the chief spacecraft with respect to the $\mathcal{F}_{LVLH}$ frame is described by the quaternion $\mathbf{q}_{c}^{LVLH} = [1,\, 0,\, 0,\, 0]$, expressed using the scalar-first convention. This unit quaternion defines the orientation of the body frame $\mathcal{F}_{c}$ relative to the reference frame $\mathcal{F}_{LVLH}$. Throughout this paper, the superscript denotes the reference frame, while the subscript denotes the body frame. Accordingly, the angular velocity of the chief spacecraft is defined as $\bm{\omega}_{c}^{LVLH} = [0,\, 0,\, 0]$.

The relative position $\mathbf{r}_{d}^{LVLH}$ and relative velocity $\mathbf{v}_{d}^{LVLH}$ between the deputy and chief spacecraft can be determined by equation \ref{eq1} and equation \ref{eq2}, respectively.
\begin{equation} \label{eq1}
    \mathbf{r}_{d}^{LVLH}=\mathbf{R}_{\text{ECI}}^{LVLH} \cdot (\mathbf{r}^{\text{ECI}}_{d} - \mathbf{r}^{\text{ECI}}_{c})
\end{equation}
\begin{equation} \label{eq2}
    \mathbf{v}_{d}^{LVLH}=\mathbf{R}_{ECI}^{LVLH} \cdot (\mathbf{v}^{\text{ECI}}_{d} - \mathbf{v}^{\text{ECI}}_{c}) - \bm{\omega}^{LVLH} \times \mathbf{r}^{LVLH}_{d}
\end{equation}
where $\mathbf{R}_{\text{ECI}}^{LVLH} \in \mathbb{R}^{3\times3}$ is the rotation matrix from the ECI frame to the LVLH frame, constructed from the radial, transverse, and orbital normal unit vectors. The angular velocity term $\bm{\omega}^{LVLH}$ accounts for the rotation of the LVLH frame, thereby yielding the relative velocity expressed in a non-inertial reference frame. The initial position of the chief spacecraft is determined by its Classical Orbital Elements (COEs), while the initial relative position of the deputy spacecraft is randomly generated such that the relative distance satisfies $D_{d}^{c} \in [75, 125]$.

The initial attitude and angular velocity of the deputy spacecraft with respect to the LVLH frame are randomly initialized, where $\| \mathbf{q}_{d}^{LVLH} \| = 1$ and $\bm{\omega}_{d}^{LVLH} \in [0,1)^3$. Since the chief spacecraft body frame $\mathcal{F}_c$ is coincident with the LVLH frame, the relative attitude and relative angular velocity of the deputy spacecraft with respect to the chief spacecraft can be directly obtained by equation \ref{eq3} and equation \ref{eq4}.

\begin{equation} \label{eq3}
    \mathbf{q}_{d}^{c} = (\mathbf{q}_{c}^{LVLH})^{-1} \otimes \mathbf{q}_{d}^{LVLH}
\end{equation}
\begin{equation}\label{eq4}
    \bm{\omega}_{d}^{c} = \bm{\omega}_{d}^{LVLH} - \bm{\omega}_{c}^{LVLH}
\end{equation}

As the body frame $\mathcal{F}_{c}$ is coincident with the LVLH frame, it can be considered that $\mathbf{q}_{d}^{c}=\mathbf{q}_{d}^{LVLH}$ and $\bm{\omega}_{d}^{c}=\bm{\omega}_{d}^{LVLH}$.

Since a Keplerian circular orbit is assumed and the relative distance between the spacecraft is sufficiently small, the translational relative motion is modeled using the linear Clohessy--Wiltshire (CW) equations \cite{clohessy1960terminal} to describe the evolution of the relative position and velocity:
\begin{equation}\label{eq5}
  \begin{split}
    &\ddot{x}_{d}^{c}+2n_{0}\dot{y}_{d}^{c} = f_{t, x}\\
    &\ddot{y}_{d}^{c}-2n_{0} \dot{x}_{d}^{c} - 3 n_{0}^{2}y_{d}^{c} = f_{t, y}\\
    &\ddot{z}_{d}^{c}+n_{0}^{2}z_{d}^{c}=f_{t, z}
   \end{split}
\end{equation}
where $f_{t, x}$, $f_{t, y}$, and $f_{t, z}$ denote the normalized thrust accelerations applied to the deputy spacecraft along the LVLH axes at each time step. The parameter $n_{0}=\sqrt{\mu / r^{3}} \approx 9.720 \times 10^{-4}\,\mathrm{rad/s}$ represents the orbital angular velocity, where $\mu$ is the Earth's gravitational parameter and $r$ is the orbital radius. The rotational dynamics of the deputy spacecraft are described by the quaternion kinematics and rigid-body rotational equations given in \eqref{eq6} and \eqref{eq7}, respectively:
\begin{equation}\label{eq6}
    \dot{\mathbf{q}}_{d}^{c} = \frac{1}{2}\Omega(\boldsymbol{\omega}_{d}^{c}) \mathbf{q}_{d}^{c}
\end{equation}
\begin{equation}\label{eq7}
    \dot{\bm{\omega}}_{d}^{d}
    = \textbf{J}_{d}^{-1}
    \bigl(
    \bm{\tau}_{d}
    - \bm{\omega}_{d}^{d} \times (\textbf{J}_{d}\bm{\omega}_{d}^{d})
    \bigr)
\end{equation}
where $\Omega(\cdot)$ denotes the quaternion kinematic matrix constructed from the relative angular velocity $\bm{\omega}_{d}^{c}$, and $\textbf{J}_{d}$ represents the inertia matrix of the deputy spacecraft. Since the angular velocity of the chief spacecraft in the LVLH frame is set to zero, i.e., $\bm{\omega}_{c}^{LVLH} = \mathbf{0}$, the relative angular velocity $\bm{\omega}_{d}^{c}$ is numerically equivalent to the deputy body angular velocity $\bm{\omega}_{d}^{d}$. The control torque $\bm{\tau}_{d}$ is applied to the deputy spacecraft and is defined in the body frame $\mathcal{F}_{d}$. Therefore, when coupling with translational dynamics expressed in the LVLH frame, the applied torques are required to be transformed accordingly.

State propagation is performed using the fourth-order Runge–Kutta (RK4) integration scheme to ensure a good balance between computational efficiency and numerical accuracy. At each time step, a total of 13 state variables are recorded, including the relative position $\left[r_{rel,x},\, r_{rel,y},\, r_{rel,z}\right]$, relative velocity $\left[v_{rel,x},\, v_{rel,y},\, v_{rel,z}\right]$, relative attitude represented by a quaternion $\left[q_{rel,w},\, q_{rel,x},\, q_{rel,y},\, q_{rel,z}\right]$, and relative angular velocity $\left[\omega_{rel,x},\, \omega_{rel,y},\, \omega_{rel,z}\right]$. Since the chief spacecraft undergoes free-floating orbital motion, its absolute position, velocity, and attitude states in the $\mathcal{F}_{ECI}$ frame can be directly obtained from the orbital dynamics. By combining the absolute state of the chief spacecraft with the relative states provided by expert demonstrations, the absolute states of the deputy spacecraft in the $\mathcal{F}_{ECI}$ frame can be uniquely determined.

\subsection{Expert Demonstrations}
In many existing studies, expert demonstrations for spacecraft rendezvous and docking tasks are commonly collected through teleoperation. However, teleoperated demonstrations are often contaminated by oscillatory behaviors induced by human reaction delays and control inaccuracies, which can result in unnecessary fuel consumption and suboptimal control performance. To overcome these limitations, an MPC framework is employed to generate expert demonstrations. By explicitly accounting for system dynamics and control constraints, MPC produces smooth, dynamically consistent, and fuel-efficient control trajectories, thereby providing high-quality expert data for imitation learning. Specifically for spacecraft rendezvous and docking, a nonlinear MPC is adopted to solve the corresponding optimal control problem. The target state $\hat{\mathbf{s}}$ is defined as
\begin{equation} \label{eq10}
    \hat{\mathbf{s}} = 
    \begin{bmatrix} 
        \hat{\mathbf{r}} \\
        \mathbf{0} \\
        \hat{\mathbf{q}} \\ 
        \mathbf{0}
    \end{bmatrix}
\end{equation}
where $\hat{\mathbf{r}}$ and $\hat{\mathbf{q}}$ denote the desired relative position and attitude at the docking configuration, while zero relative velocity and angular velocity are enforced at the terminal state.

Considering that the coordinate frame of the docking port is not coincident with $\mathcal{F}_{c}$, $\hat{\mathbf{r}}$ and $\hat{\mathbf{q}}$ are thus defined as non-zero vectors. The overall error vector is defined by
\begin{equation} \label{eq11}
    \textbf{e}_{t} = \begin{bmatrix} 
        \mathbf{r}_{t} - \hat{\mathbf{r}}\\
        \mathbf{v}_{t}\\ 
        \mathbf{e}_{q,t}\\
        \bm{\omega}_{t}
    \end{bmatrix}
\end{equation}
where $\mathbf{e}_{q,t} = \hat{\mathbf{q}}^{-1} \otimes \mathbf{q}_{t}$. The cost function is defined in terms of the state error, control effort, and terminal error, as shown in
\begin{equation}\label{eq12}
    \textit{\textbf{J}} = \sum_{t=0}^{N_{p}-1}
    \left(
    \textbf{e}_{t}^{T}\textbf{Q}\textbf{e}_{t}
    +
    \textbf{u}_{t}^{T}\textbf{R}\textbf{u}_{t}
    \right)
    +
    \textbf{e}_{N_{p}}^{T}\textbf{P}\textbf{e}_{N_{p}}
\end{equation}
Here, $\textbf{Q}$, $\textbf{R}$, and $\textbf{P}$ are the weighting matrices for the state error, control effort, and terminal error, respectively.

The optimization method adopted for this task is Sequential Quadratic Programming (SQP). The resulting optimization problem is formulated as follows:
\begin{equation}
\begin{aligned}
\min_{\mathbf{u}_{0:N_c-1}} \quad & J(\textbf{s}_0, \mathbf{u}_{0:N_c-1}) \\[2pt]
\text{s.t.} \quad 
& \textbf{s}_{t+1} = f_d(\textbf{s}_t, \mathbf{u}_t), 
\quad t = 0,\dots,N_p-1, \\[2pt]
& \mathbf{u}_{\min} \le \mathbf{u}_k \le \mathbf{u}_{\max},
\quad k = 0,\dots,N_c-1 .
\end{aligned}
\end{equation}
Here, $N_{p}$ and $N_{c}$ represent the prediction horizon and control horizon, respectively, and $f_d$ denotes the discrete-time state transition function defined by the system dynamics. To better reflect real-world operating conditions, random noise is further injected into the expert demonstrations to emulate unknown disturbances and sensor uncertainties.

\begin{figure}[!htbp]
    \centering
    \includegraphics[scale=0.85, trim= 60 0 40 0]{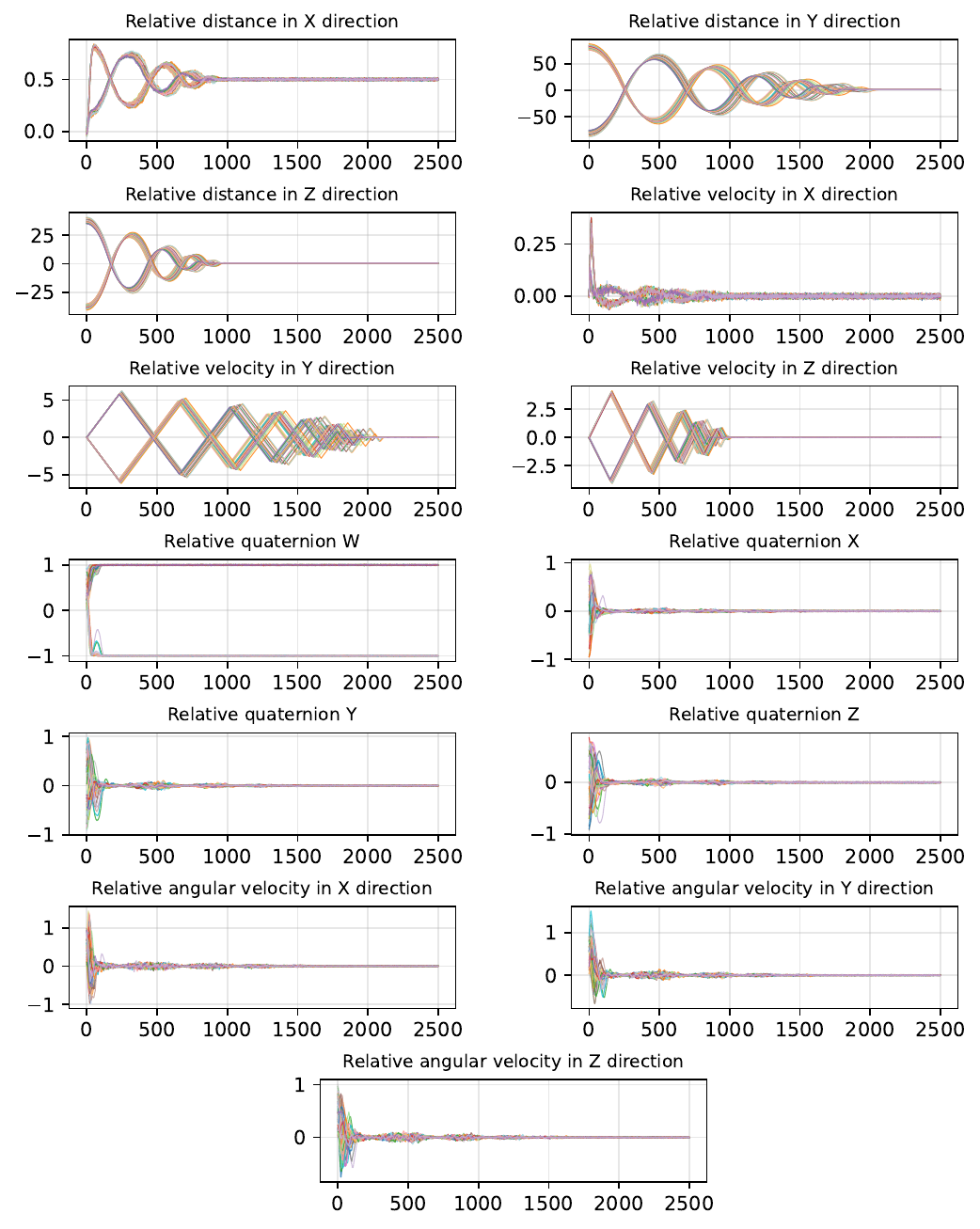}
    \caption{Distribution of expert demonstrations}
    \label{fig:expert_demo}
\end{figure} 

A total of 50 expert trajectories are collected, each with an episode length of 2500 time steps to ensure full convergence of both translational and rotational states. The distribution of the expert demonstrations is illustrated in Fig. \ref{fig:expert_demo}. When unknown disturbances are introduced, the expert trajectories exhibit noticeable deviations from smooth trajectories, particularly during the terminal phase of the rendezvous and docking process.

The initial conditions span both positive and negative values. This design choice prevents the proposed algorithm from learning a trivial state-to-action mapping and instead encourages it to capture the global convergence behavior toward the desired terminal state. It is also observed that the scalar component of the relative quaternion converges to either $1$ or $-1$, which reflects the inherent sign ambiguity of quaternion representations. Since both values correspond to the same physical attitude, this ambiguity does not affect the final docking configuration.

\subsection{IL-SRD Method}

Spacecraft rendezvous and docking is a long-horizon control task in which control actions must remain temporally consistent to ensure stable convergence. In such scenarios, even small deviations in early predicted actions can accumulate over time and significantly degrade closed-loop performance. Conventional imitation learning approaches typically learn a direct state-to-action mapping at each timestep and do not explicitly model temporal correlations across the control horizon. As a result, they often produce inconsistent or oscillatory action sequences during long-term execution. To address this limitation, we propose the IL-SRD method, which explicitly models temporal dependencies in expert demonstrations. The overall architecture is illustrated in Fig. \ref{fig:network}.

We first employ a Variational Auto-Encoder (VAE) to capture the latent representations underlying the expert demonstrations. Specifically, the current system state and the subsequent expert action sequence are jointly provided as inputs to the VAE. To facilitate effective representation learning and feature alignment, the state and action inputs are first mapped into a shared high-dimensional latent space via separate projection networks:
\begin{equation}
    \begin{aligned}
        \textbf{Embed}_{s} &= \psi_{s}(\textbf{s}),\\
        \textbf{Embed}_{a_t} &= \psi_{a}(\hat{\textbf{a}}_{t}),
    \end{aligned}
\end{equation}
where $\psi_{s}$ and $\psi_{a}$ denote the projection networks used to preprocess the state and expert action sequence, respectively, with $\hat{\textbf{a}}_{t} \in \mathcal{A}^{\text{exp}}_{t:t+n}$, and $\mathcal{A}^{\text{exp}}$ representing the expert action sequence.

The input to the VAE is then formulated as follows:
\begin{equation}
    \textbf{E} = \left([\textbf{Embed}_{s}, \textbf{Embed}_{a_{1}}, \ldots, \textbf{Embed}_{a_{n}}] + \textbf{P}_{\text{VAE}}\right)
    \in \mathbb{R}^{(n+1)\times d},
\end{equation}
where $d$ denotes the shared embedding dimension used in the Transformer encoder--decoder architecture, $n$ represents the length of the expert action sequence, and $\textbf{P}_{\text{VAE}}$ denotes the corresponding positional encoding.

Unlike conventional imitation learning methods that predict a single control action at each timestep, the proposed algorithm predicts a sequence of control actions of length $n$. This sequence-based prediction strategy enables the model to capture the latent structure and temporal correlations underlying expert control behaviors. Moreover, by operating at a coarser temporal resolution with fewer prediction steps, the proposed approach effectively mitigates error accumulation during long-horizon execution.

The Transformer captures temporal dependencies primarily through the self-attention mechanism, which operates on a sequence of embedded tokens augmented with positional encodings. For a given attention head, the query matrix $\textbf{Q}_{enc} = \textbf{E}\textbf{W}^{Q}_{enc} \in \mathbb{R}^{N_{q}\times d_{k}}$, key matrix $\textbf{K}_{enc} = \textbf{E}\textbf{W}^{K}_{enc} \in \mathbb{R}^{N_{k}\times d_{k}}$, and value matrix $\textbf{V}_{enc} = \textbf{E}\textbf{W}^{V}_{enc} \in \mathbb{R}^{N_{k}\times d_{v}}$ are obtained via learned linear projections, where $N_q$ and $N_k$ denote the numbers of query and key/value tokens, respectively.

The attention weights are computed using the scaled dot-product operation, allowing each query to selectively attend to relevant temporal tokens. For notational simplicity, the subscripts are omitted in the following formulation. The multi-head attention (MHA) mechanism is defined as:
\begin{equation}
    \begin{aligned}
        \text{Attn}(\textbf{Q}, \textbf{K}, \textbf{V}) &= 
        \text{softmax}\!\left(\frac{\textbf{Q}\textbf{K}^{T}}{\sqrt{d_{k}}}\right)\textbf{V},\\
        \text{head}_{h} = 
        \text{Attn}&(\textbf{Q}\textbf{W}^{Q}_{h}, \textbf{K}\textbf{W}^{K}_{h}, \textbf{V}\textbf{W}^{V}_{h}),\\
        \textbf{MHA}(\textbf{Q}, \textbf{K}, \textbf{V}) = 
        &\text{Concat}(\text{head}_{1}, \ldots, \text{head}_{H})\textbf{W}^{O},
    \end{aligned}
\end{equation}
where $\textbf{W}^{Q}_{h}$, $\textbf{W}^{K}_{h}$, and $\textbf{W}^{V}_{h}$ are the learnable projection matrices for the query, key, and value in the $h$-th attention head, respectively. The matrix $\textbf{W}^{O}$ denotes the output projection matrix that aggregates the concatenated multi-head features.

The output of the MHA module is further processed through residual connections, feed-forward networks, and layer normalization, yielding the final encoder output $\text{Encoder}(\textbf{E}) = \textbf{H}_{enc}$. The encoder output is then linearly projected to obtain the mean vector $\bm{\mu}$ and the log-variance vector $\log \bm{\sigma}^{2}$. The mean $\bm{\mu}$ encodes the latent representation of the input, while $\bm{\sigma}$ characterizes the uncertainty associated with this representation. Together, these two values define an approximate posterior distribution over the latent space.

To enable backpropagation through the stochastic sampling process, the latent variable $\textbf{z}$ is obtained using reparameterization:
\begin{equation} \label{eq13}
    \textbf{z} = \bm{\mu} + \bm{\sigma} \odot \bm{\epsilon}, 
    \quad \bm{\epsilon} \sim \mathcal{N}(\mathbf{0}, \mathbf{I}),
\end{equation}
where $\mathbf{I}$ denotes the identity covariance matrix, implying that each dimension of $\bm{\epsilon}$ is independently sampled from a standard normal distribution. The latent variable $\textbf{z}$ captures the underlying structure of expert demonstrations and facilitates the decoder in generating physically consistent and stable control predictions.

During evaluation, since future expert action sequences are not available, $\textbf{z}$ is set to a zero vector. The decoder memory is then constructed as:
\begin{equation}
    \textbf{M} = [\phi_{z}(\textbf{z}), \psi_{s}(\textbf{s}_{t})] \in \mathbb{R}^{2 \times d},
\end{equation}
where $\phi_{z}$ and $\psi_{s}$ denote the projection networks for the latent variable and the current state feature, respectively.

To further enhance the physical consistency and temporal coherence of the predicted action sequence, we propose an ADT mechanism, which embeds state-dependent structural priors into the decoding process. Specifically, the current spacecraft state is replicated along the prediction horizon and embedded as the decoder target input. This anchoring strategy ensures that each predicted action is explicitly conditioned on the same physical state reference, rather than relying solely on positional embeddings or autoregressive action tokens. As a result, the decoder predicts control sequences as structured deviations from the current state, instead of producing actions in an unconstrained latent space. In addition, the semantic consistency between the decoder target and the encoder memory mitigates representation mismatch during backpropagation, leading to more stable and efficient convergence. The ADT is defined as follows:

\begin{equation}
    \textbf{Y}_{t} = \left([\psi_{s}(\textbf{s}_{t}), ..., \psi_{s}(\textbf{s}_{t})] + \textbf{P}_{ADT}\right)\in \mathbb{R}^{n\times d}
\end{equation}

Similarly, both memory and the ADT are fed into MHA such that $\textbf{Q}_{dec} = \textbf{Y} \textbf{W}^{Q}_{dec} \in \mathbb{R}^{N_{q}\times d_{k}}$, key matrix $\textbf{K}_{dec} = \textbf{M} \textbf{W}^{K}_{dec} \in \mathbb{R}^{N_{k}\times d_{k}}$ and value matrix $\textbf{V}_{dec} = \textbf{M} \textbf{W}^{V}_{dec} \in \mathbb{R}^{N_{k}\times d_{v}}$. The decoder output is subsequently processed through the same post-processing layers as those used in the VAE, and finally passed through an action projection layer to produce the predicted action sequence:

\begin{equation}
    \mathcal{A}^{pred}_{t:t+n} = \text{Decoder}(\textbf{Y}, \textbf{M})
\end{equation}

\begin{figure}[ht]
    \centering
    \includegraphics[width=0.8\linewidth, trim=10 10 10 10]{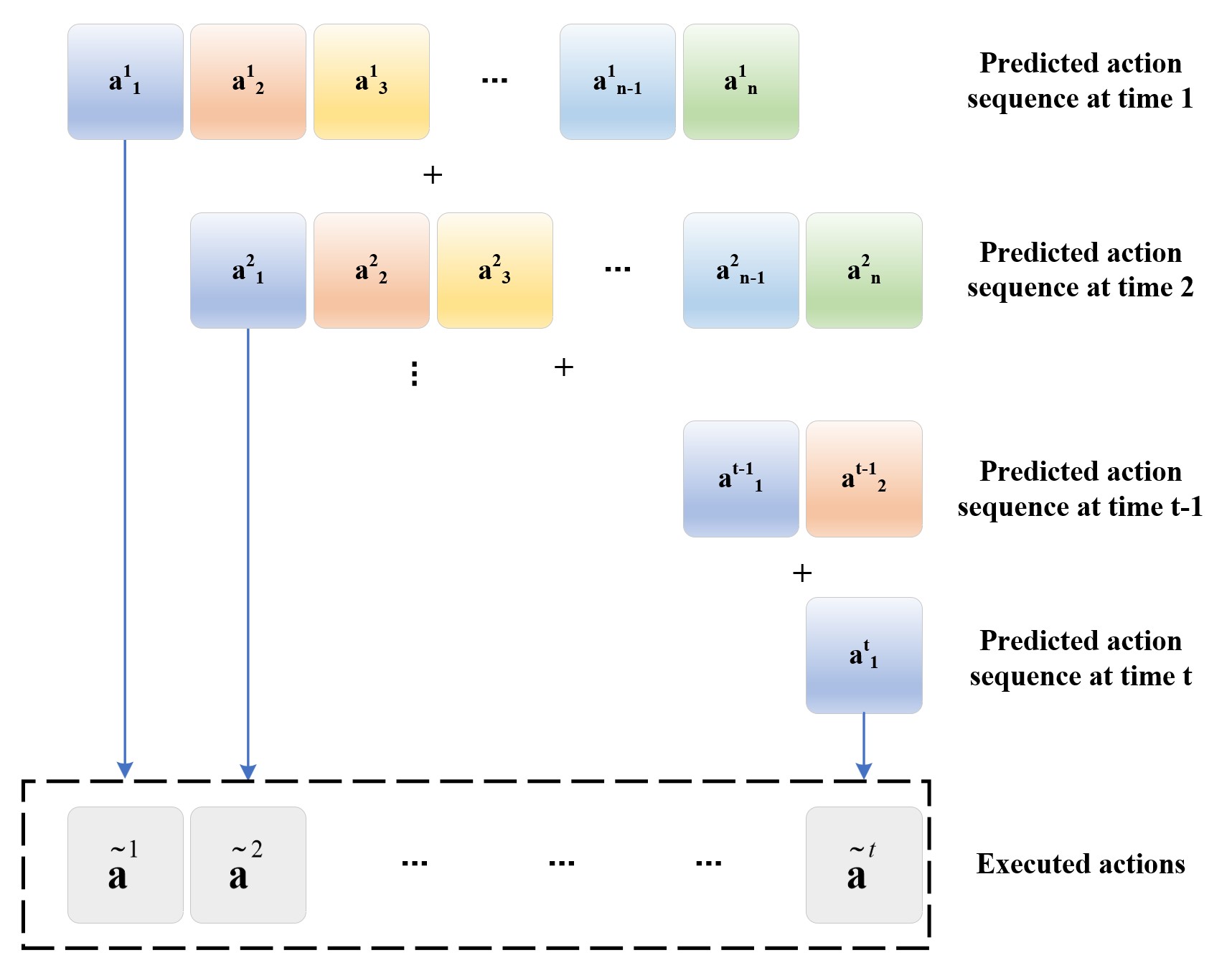}
    \caption{Demonstration of temporal aggregation}
    \label{fig:ac}
\end{figure}
where $\mathcal{A}^{pred}_{t:t+n}=[\textbf{a}^{t}_{1}, \dots, \textbf{a}^{t}_{n}]$, indicates that this action sequence is predicted at time $t$ with the sequence length as $n$. The superscript of each action denotes the time step at which the action sequence is predicted, while the subscript indicates the index of the action within the predicted action sequence. To further eliminate the effects of accumulated predicted errors and unknown disturbances induced in the expert demonstrations, we introduce the temporal aggregation mechanism illustrated is Fig.\ref{fig:ac}. The final action for each timestep $\tilde{\textbf{a}}^{t}$ is then defined as follows:
\begin{equation}
    \begin{aligned}
        w_i &= \exp\left( -\kappa \cdot i \right), \quad i = 1,\dots,n, \\
        \tilde{w}_i &= \frac{w_i}{\sum_{j=1}^{n} w_{j}}, \\
        \tilde{\textbf{a}}^{t} &= \displaystyle\sum_{i=1}^{n} \tilde{w}_i \cdot \textbf{a}^{t-n+i}_{n-i+1}, \text{if} \ t \geq n,\\
        \tilde{\textbf{a}}^{t} &= \displaystyle\sum_{i=1}^{t} \tilde{w}_i \cdot \textbf{a}^{i}_{t-i+1}, \text{else}
    \end{aligned}
\end{equation}

By utilizing temporal aggregation, the finalized action at each timestep $\tilde{\textbf{a}}^{t}$ is a weighted sum of all the previous predicted actions at this timestep. 

There are two parts of loss function proposed to facilitate convergence of IL-SRD algorithm. The first part evaluates the deviation between the predicted and the expert action sequences. Since the action sequences contain physical-scale variables (the attitude information which is presented in the form of quaternions), therefore the deviation loss is demonstrated by equation \ref{eq8}.

\begin{equation} \label{eq8}
\begin{aligned}
l_{r} &= \frac{1}{M} \sum_{1}^{M}\text{MSE}(\textbf{r}^{t}_{i}, \hat{\textbf{r}})\\
l_{v} &= \frac{1}{M} \sum_{1}^{M}\text{MSE}(\textbf{v}^{t}_{i}, \hat{\textbf{v}})\\
l_{q} &= \frac{1}{M} \sum_{1}^{M}\alpha(\textbf{q}^{t}_{i}, \hat{\textbf{q}})^{2} \\
l_{\omega} &= \frac{1}{M} \sum_{1}^{M}\text{MSE}(\bm{\omega}^{t}_{i}, \hat{\bm{\omega}})\\
\end{aligned}
\end{equation}
in which $M$ represents the batch size, $\alpha(\textbf{q}^{t}_{i}, \hat{\textbf{q}}) = 2\cdot \text{arccos}(\left| \text{Re}(\textbf{q}^{t}_{i} \otimes \hat{\textbf{q}}^{-1})\right|)$, and $\text{Re}(\cdot)$ means to take the real part of the quaternion, the predicted and expert action sequence follow the same notation as defined above. All expert demonstrations are normalized to mitigate the effects of heterogeneous data magnitudes and enhance training stability. Based on the formulations above, losses are jointly computed on both the normalized and de-normalized representations and combined in a weighted manner. This strategy leverages the numerical stability of normalized learning while simultaneously enforcing physical-scale consistency, thereby effectively reducing the discrepancy between expert demonstrations and predicted trajectories in both feature and physical spaces.

The second term of the loss function corresponds to the Kullback-Leibler (KL) divergence, which measures the discrepancy between the approximate posterior distribution produced by the VAE and a unit Gaussian prior $\mathcal{N}(0, \mathbf{I})$. This regularization term encourages the learned latent space to remain smooth, continuous, and well-structured. The KL loss function is presented as follows:
\begin{equation}
    l_{KL} = mean(\sum -\frac{1}{2} \cdot (1 + \log\,\sigma^{2} - \mu^{2}-\sigma^{2}))
\end{equation}
in which $\mu$ and $\sigma$ are the processed output of the VAE, the same as those used in equation \ref{eq13}. So the final loss function is demonstrated as such:
\begin{equation}
    \mathcal{L} = \lambda_{r} l_{r} + \lambda_{v} l_{v} + \lambda_{q} l_{q} + \lambda_{\omega} l_{\omega} + \lambda_{KL} l_{KL}
\end{equation}
where $\lambda_{r}, \lambda_{v}, \lambda_{q}, \lambda_{\omega}, \lambda_{KL}$ are the respective weights.

\section{Experimental Results} \label{sec: simulation}
\subsection{Experimental Setting}
We compare against five baseline methods, including a PID-based method, a MPC-based method, two DRL-based methods (i.e. D4PG \cite{hovell2021deep} and SAC \cite{shao2025asymmetric}) and a vanilla BC method. 

For the PID-based and MPC-based methods, the relative dynamic model established to collect expert trajectories is used. Maximum unit thrust limits are set to $F_{thrust}=[\pm 0.2, \pm 0.2, \pm 0.2]\ N/kg$, and the max torques applied on each axis are set to $\tau = [\pm 8, \pm 8, \pm 8] \ N\cdot m$. 

The MPC-based method is identical to the controller used to generate expert demonstrations. By incorporating it as a comparative algorithm, it can be shown how the proposed algorithm is able to capture the temporal features of the expert demonstration and provide a direct evaluation of the performance. 

As for the PID-based control scheme, we design a dual-loop independent PID architecture that decouples translational and rotational control. In addition, a second-order low-pass filter combined with composite differentiation, together with active damping techniques, is employed to attenuate attitude and angular-velocity oscillations.

The D4PG algorithm \cite{hovell2021deep} is a DRL–based spacecraft rendezvous and docking control method built upon the DDPG (Deep Deterministic Policy Gradient) framework \cite{lillicrap2015continuous}. Unlike classical DDPG, D4PG uses multiple agents to improve sample efficiency and the critic network predicts a full value distribution rather than a single expected return. By modeling the return distribution, D4PG can more effectively
capture environmental uncertainties, leading to more accurate value estimation and improved policy performance.

Our previous work \cite{shao2025asymmetric} employs the SAC algorithm \cite{haarnoja2018soft}, augmented with privileged learning, to perform spacecraft approaching maneuver. In this framework, the actor network receives perturbed or partially observed information, while the critic has access to privileged states during training, enabling more robust policy learning under informational disturbances. 

The dynamic models adopted in \cite{hovell2021deep} and \cite{shao2025asymmetric} differ from the one used in this paper. Specifically, \cite{hovell2021deep} employs a simplified 2-D model that includes only the translational dynamics in the $x-y$ plane along with yaw rotation. In contrast, \cite{shao2025asymmetric} considers only translational motion and completely omits rotational dynamics. To ensure a fair comparison with the proposed method, both DRL-based approaches are extended to 6-DOF control and re-trained from scratch under identical conditions.

For the vanilla BC algorithm, a 5-layer MLP (Multi-Layer Perceptron) network is used instead of a Transformer-based network for imitation learning. Unlike the proposed IL-SRD algorithm, which predicts a sequence of actions, vanilla BC predicts only a single action at each timestep, and thus does not employ the proposed temporal aggregation mechanism. Both approaches use the same normalization and post-processing techniques to ensure that the output actions remain within their limits.

Five evaluation metrics are considered, including the Convergence Step (CS), Stepwise Energy Consumption (SEC), the Average Terminal Translational Precision (ATTP), Average Terminal Rotational Precision (ATRP), and Episodic Stepwise Reward (ESR). 

CS is chosen to be the step which leads the start of a consecutive $20$ steps that has the minimum mean error compared to the final state to ensure that there will be no aggressive changes for the following steps.

Certain control methods directly output control actions, such as PID, MPC, SAC and D4PG. For such algorithms, the output control actions are used to calculate SEC metric. However, for algorithms such as IL-SRD and Vanilla BC, the energy is calculated based on consecutive state transitions. To avoid noise amplification, a central difference scheme is employed to compute the linear acceleration and angular acceleration:
\begin{equation}
    \begin{aligned}
        \dot{\textbf{v}}_{t} \approx & \frac{\textbf{v}_{t+1} - \textbf{v}_{t-1}}{2\Delta t}\\
        \dot{\bm{\omega}}_{t} \approx & \frac{\bm{\omega}_{t+1} - \bm{\omega}_{t-1}}{2\Delta t}
    \end{aligned}
\end{equation}
$\Delta t$ is the control interval and is set to 0.1s in this paper. Since the translational dynamics are formulated based on linear CW equations, and the gravitational acceleration $\textbf{acc}_{G}$ is included and can be derived from equation \ref{eq5}:

\begin{equation}
    \textbf{acc}_{G, t} = \left[ \begin{matrix}
    -2n_{0}\dot{y}_{t}\\
    2n_{0}\dot{x}_{t}+3n_{0}^{2}y_{t}\\
    -n_{0}^{2}z_{t}
    \end{matrix}\right]
\end{equation}

The translational acceleration $\textbf{acc}_{thr}$ induced by the applied thrust is expressed as:
\begin{equation}
    \textbf{acc}_{thr, t} = \dot{\textbf{v}}_{t} - \textbf{acc}_{G, t}
\end{equation}

Correspondingly, the resulting thrust force is obtained from the translational dynamics as:
\begin{equation}
    \textbf{F}_{t} = m_{d} \cdot \textbf{acc}_{thr, t}
\end{equation}

The control torque can be inferred from the rotational dynamics as:
\begin{equation}
    \bm{\tau}_{t} = \textbf{J}_{d}\dot{\bm{\omega}}_{t} + \bm{\omega}_{t} \times (\textbf{J}_{d}\bm{\omega}_{t})
\end{equation}

Based on the resulting thrust and torque, the overall energy consumption is evaluated as:
\begin{equation}
    \begin{aligned}
        \mathbf{W}_{\textbf{F}} = &\sum_{t} ||\textbf{F}_{t}\odot (\textbf{r}_{t+1} - \textbf{r}_{t})||_{1}\\
        \mathbf{W}_{\bm{\tau}} =  &\sum_{t} ||\bm{\tau}_{t}\odot \Delta \bm{\theta}_{t}||_{1}
    \end{aligned}
\end{equation}
where $\Delta \bm{\theta}_{t}$ is the incremental attitude displacement. SEC is computed based on the episode length for each method. 

ESR is obtained by accumulating the reward over an entire epoch and then averaging it over the episode length. ESR is then calculated as follows:
\begin{equation}
            \text{ESR} = \frac{1}{N_{steps}} \sum^{N_{steps}}_{t=1} -(\lVert\textbf{r}_{t}- \hat{\textbf{r}}\rVert_{2} + \lVert\textbf{v}_{t}- \hat{\textbf{v}}\rVert_{2}
     + \alpha(\textbf{q}_{t}, \hat{\textbf{q}})^{2} + \lVert\omega_{t}- \hat{\bm{\omega}}\rVert_{2})
\end{equation} 
where $\alpha(\textbf{q}_{t}, \textbf{q}_{t})^{2}$ is the same as the one used in equation \ref{eq8}.

ATTP and ATRP are to determine the control precision of the final state of each episode. Since this paper targets a 6-DOF spacecraft rendezvous and docking control task, the terminal translational precision and rotational precision for a single evaluation episode are determined separately:
\begin{equation}
    \begin{aligned}
           &\text{TTP} = \lVert\textbf{r}_{final} - \hat{\textbf{r}}\rVert_{2} + \lVert\textbf{v}_{final} - \hat{\textbf{v}}\rVert_{2}\\
           &\text{TRP} = \alpha(\textbf{q}_{final}, \hat{\textbf{q}})^{2} + \lVert\omega_{final} - \hat{\bm{\omega}}\rVert_{2}
    \end{aligned}
\end{equation}

All results are averaged over 5 experiments with 5 same initial states. The total evaluation step for each experiment is 2500 steps. The hyperparameters of the proposed IL-SRD algorithm are summarized in Table \ref{table1}. 
\begin{table}[!hb]
    \centering
    \caption{Hyper-parameters of the proposed IL-SRD algorithm}
    \label{table1}
    \begin{tabular}{cc}
    \hline
    Hyperparameters & Value \\ \hline
    Training Epoch  & 400   \\
    Batch Size      & 256   \\
    Sequence Length & 500   \\
    Num. of Heads   & 4     \\
    Encoder Layer   & 3     \\
    Decoder Layer   & 4     \\
    Learning Rate   & $7\times 10^{-4}$  \\
    Weight Decay    & $5\times 10^{-5}$  \\ 
    Optimizer       & AdamW\\ \hline
    \end{tabular}
\end{table}

\subsection{Performance Comparisons}

Table \ref{table2} summarizes the experimental results obtained using different control methods. The proposed algorithm outperforms all other control methods in attitude regulation and achieves performance comparable to MPC while exhibiting lower energy consumption. Moreover, IL-SRD attains the fewest convergence steps among all evaluated methods that generate an accurate rendezvous and docking process, indicating that under identical conditions it enables significantly more efficient rendezvous and docking control. A lower ESR correlates with faster convergence and smoother terminal behavior, indicating that IL-SRD achieves a smaller effective CS.

\begin{table}[!htbp]
\centering
\caption{Experimental results of different control methods}
\resizebox{\textwidth}{!}{%
\begin{tabular}{cccccc} \hline
Metrics    & CS                         & SEC   & ATTP   & ATRP  & ESR   \\ \hline
PID        & 2403.6 $\pm$ 71.0          & 1.287 $\pm$ 0.122 & 2.051 $\pm$ 0.463 & 0.011 $\pm$ 0.003 & -0.304 $\pm$ 0.494 \\
SAC        & NA                         & 19.026 $\pm$ 0.088 & 190.969 $\pm$ 0.279 & 5.118 $\pm$ 3.114 & -1.810 $\pm$ 0.552 \\
D4PG       & NA                         &  24.820 $\pm$ 2.073& 116.233 $\pm$ 55.537  & 9.128 $\pm$ 3.050 & -1.158 $\pm$ 0.647 \\
Vanilla BC & 84.6 $\pm$ 35.7           & \textbf{0.194 $\pm$ 0.125} & 39.134 $\pm$ 5.532  & 5.708 $\pm$ 3.027& -1.401 $\pm$ 0.097 \\
MPC        & 2304.4 $\pm$ 103.5         & 5.274 $\pm$ 0.425     & \textbf{0}       & 0       & -0.266 $\pm$ 0.339      \\
IL-SRD      & \textbf{2152.2 $\pm$ 80.0} & 4.166 $\pm$ 0.269 & 1.229 $\pm$ 0.185  & \textbf{0} & \textbf{-0.266 $\pm$ 0.309} \\ \hline
\end{tabular}%
}
\label{table2}
\end{table}

IL-SRD demonstrates outstanding overall performance; however, its ATTP remains inferior to that achieved by MPC. This limitation primarily arises from the characteristics of the expert demonstrations under simulated realistic conditions. In order to approximate real-world onboard environments, unknown sensor errors are injected during data generation, which introduce small oscillatory patterns into the expert trajectories. These sensor-induced fluctuations are inevitably captured by the proposed imitation learning framework. During evaluation, the learned oscillatory tendencies lead to slight deviations between the final translational state and the desired target.

PID, on the other hand, achieves the second-lowest SEC, following Vanilla BC. This indicates that PID tends to choose less aggressive actions compared to MPC which results in longer CS. Both ATTP and ATRP for PID are worse than IL-SRD, this is because PID relies on instantaneous error feedback and lacks the ability to explicitly account for long-horizon dynamics. As a result, the PID controller tends to exhibit small-amplitude oscillations during the final stage of the spacecraft rendezvous and docking process.

Both SAC and D4PG fail to converge and instead reach the termination conditions defined in the simulation environment. This primarily stems from the complexity of the reward function, which comprises multiple components, making it difficult for the DRL agents to effectively adjust their network parameters. In addition, the relative translational states (positions and velocities) and rotational states (attitude quaternions and angular velocities) are strongly coupled. Although \cite{shao2025asymmetric} and \cite{hovell2021deep} report successful convergence, the task considered in this paper is substantially more challenging. Under the increased task complexity, neither SAC nor D4PG is able to control the full 6-DOF spacecraft to complete the entire rendezvous and docking maneuver. The enlarged state and action spaces, together with the more intricate reward structure, exacerbate the learning difficulty. As a result, algorithms that previously converged under simpler settings become distracted by the high-dimensional coupling effects, ultimately producing unstable and highly divergent control policies.

Although Vanilla BC achieves the lowest CS, the remaining metrics indicate that it converges to an arbitrary state rather than the desired terminal state. Vanilla BC also attains the minimum SEC; however, this is primarily because, after converging to an arbitrary state, its control policy ceases to generate further control actions. As demonstrated by the results, this single step prediction tends to provide actions that are close to the initial state. 

\begin{figure}[!htbp]
    \captionsetup[sub]{
    aboveskip=0pt,    
    belowskip=0pt,    
    margin=0pt,       
    font=scriptsize   
    }
    \centering
    \subcaptionbox{}{\includegraphics[width=0.33\textwidth]{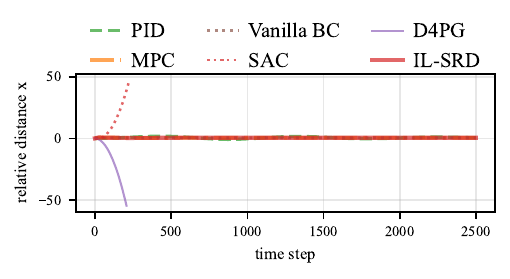}}\hfill
    \subcaptionbox{}{\includegraphics[width=0.33\textwidth]{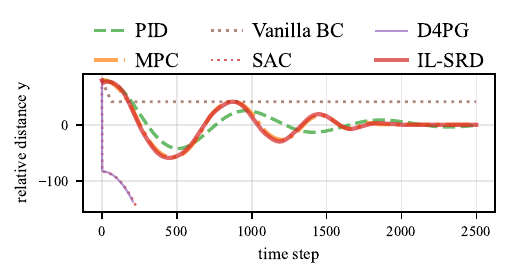}}\hfill
    \subcaptionbox{}{\includegraphics[width=0.33\textwidth]{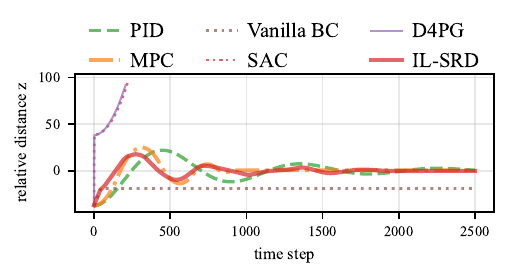}}
    \subcaptionbox{}{\includegraphics[width=0.33\textwidth]{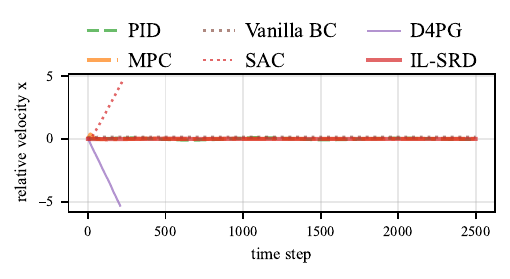}}\hfill
    \subcaptionbox{}{\includegraphics[width=0.33\textwidth]{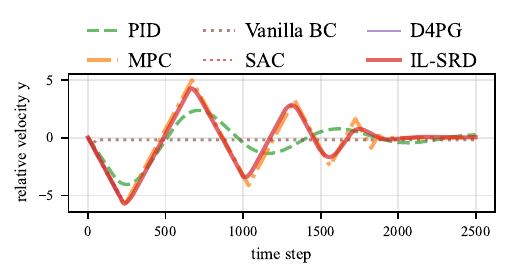}}\hfill
    \subcaptionbox{}{\includegraphics[width=0.33\textwidth]{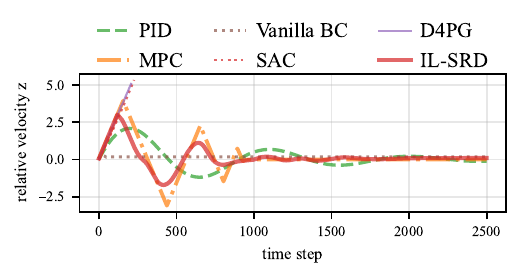}}
    \subcaptionbox{}{\includegraphics[width=0.33\textwidth]{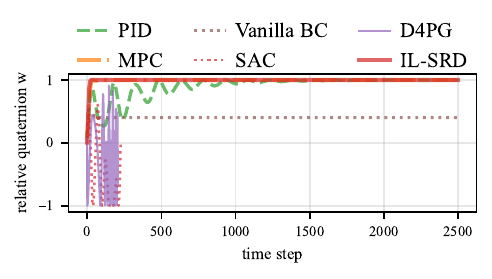}}\hfill
    \subcaptionbox{}{\includegraphics[width=0.33\textwidth]{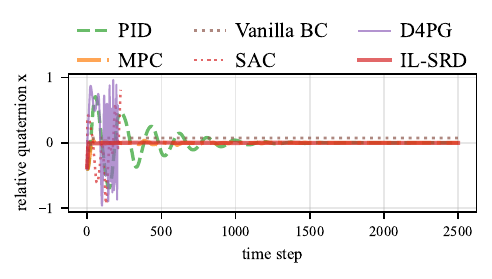}}\hfill
    \subcaptionbox{}{\includegraphics[width=0.33\textwidth]{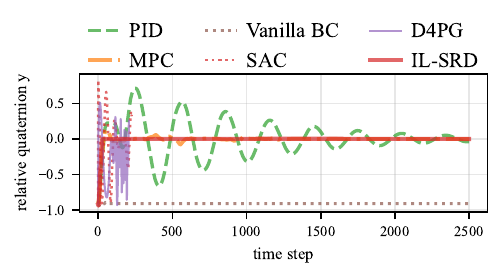}}
    \subcaptionbox{}{\includegraphics[width=0.33\textwidth]{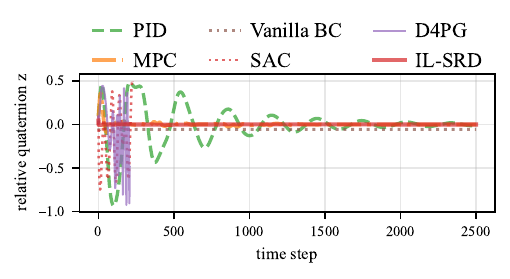}}\hfill
    \subcaptionbox{}{\includegraphics[width=0.33\textwidth]{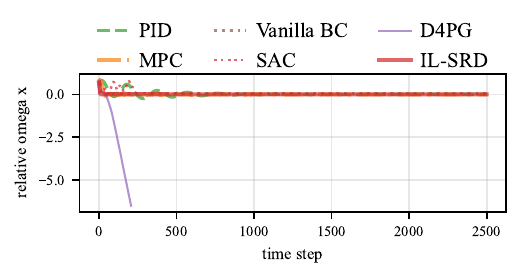}}\hfill
    \subcaptionbox{}{\includegraphics[width=0.33\textwidth]{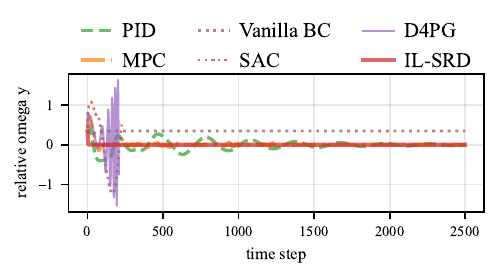}}
    \subcaptionbox{}{\includegraphics[width=0.33\textwidth]{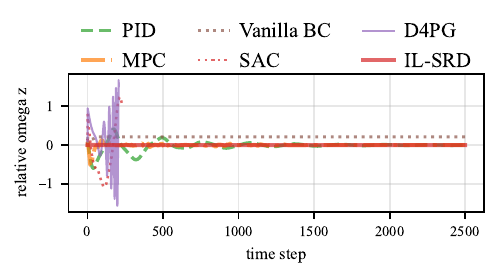}}
    
    \caption{\small Spacecraft rendezvous and docking error diagrams for different control methods, (a)--(c) are the relative distance, (d)--(e) are the relative velocity, (g)--(j) are the relative attitude represented by quaternions, (k)--(m) are the relative angular velocities}
    \label{fig:all_results_norm}
\end{figure}

Fig. \ref{fig:all_results_norm} is a demonstration of how the relative information are changed for each control method under the same initial condition. It can be seen that both MPC and IL-SRD exhibit highly consistent state evolution profiles. This consistency indicates that the proposed method effectively captures the temporal structure embedded in the expert demonstrations, enabling it to learn meaningful state–action dependencies over multiple time horizons and reproduce coherent long-horizon control behaviors.

We further evaluate the robustness of the proposed algorithm. By introducing an additional random noise to the state received for each timestep, the results are shown in Table \ref{table3}. Experimental results demonstrate that IL-SRD exhibits strong robustness to unknown disturbances: the ATTP increases by only 0.01 m, while the ATRP remains unaffected. The slight increase in SEC is attributed to more aggressive rotational maneuvers at the start of the simulation. These findings suggest that, rather than simply imitating expert trajectories, the proposed method effectively captures long-horizon temporal dependencies between the input states and the desired terminal state.

\begin{table}[!htbp]
\centering
\caption{Experimental results of different control methods under disturbed conditions}
\resizebox{\textwidth}{!}{%
\begin{tabular}{ccccc} \hline
Metrics    & SEC   & ATTP   & ATRP  & ESR   \\ \hline
PID        & \textbf{1.287 $\pm$ 0.122} & 2.102 $\pm$ 0.464 & 0.010 $\pm$ 0.003 & -0.303 $\pm$ 0.490 \\
SAC        & 63.820 $\pm$ 0 & 357.322 $\pm$ 336.380 & 12.473 $\pm$ 3.190 & -1.095 $\pm$ 0.916 \\
D4PG       &  24.191 $\pm$ 2.428 & 139.430 $\pm$ 54.953  & 10.030 $\pm$ 3.453 & -1.255 $\pm$ 0.654 \\
Vanilla BC & 149.724 $\pm$ 47.950 & 36.035 $\pm$ 1.595  & 0.187 $\pm$ 0.024 & -1.136 $\pm$ 0.192 \\
MPC        & 343.799 $\pm$ 23.355  & 285.406 $\pm$ 330.633 & 2.313 $\pm$ 3.016   & -0.583 $\pm$ 0.562      \\
IL-SRD      & 5.640 $\pm$ 0.110 & \textbf{1.238 $\pm$ 0.187}  & \textbf{0} & \textbf{-0.281 $\pm$ 0.316} \\ \hline
\end{tabular}%
}
\label{table3}
\end{table}

Under the presence of unknown disturbances, the MPC controller exhibits a noticeable degradation in performance, whereas the proposed IL-SRD framework remains largely unaffected. This discrepancy primarily stems from MPC’s intrinsic reliance on accurate state prediction over a finite horizon. When sensor noise or unknown disturbances are injected into the state measurements at each timestep, prediction errors accumulate along the optimization horizon, ultimately leading to suboptimal control actions and degraded closed-loop performance.

\begin{figure}[!htbp]
    \captionsetup[sub]{
    aboveskip=0pt,    
    belowskip=0pt,    
    margin=0pt,       
    font=scriptsize   
    }
    \centering
    \subcaptionbox{}{\includegraphics[width=0.33\textwidth]{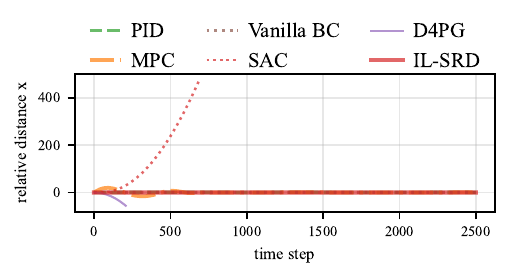}}\hfill
    \subcaptionbox{}{\includegraphics[width=0.33\textwidth]{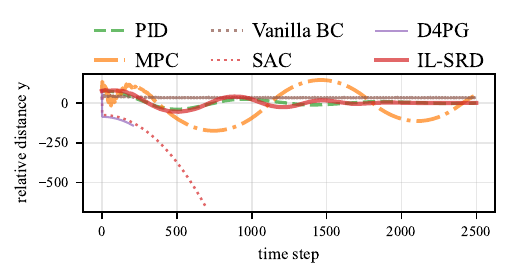}}\hfill
    \subcaptionbox{}{\includegraphics[width=0.33\textwidth]{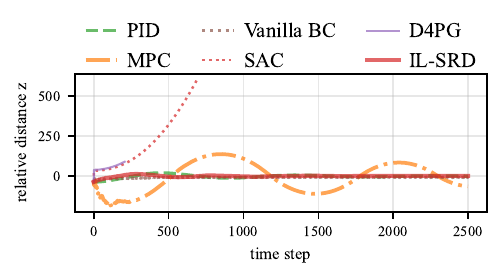}}
    \subcaptionbox{}{\includegraphics[width=0.33\textwidth]{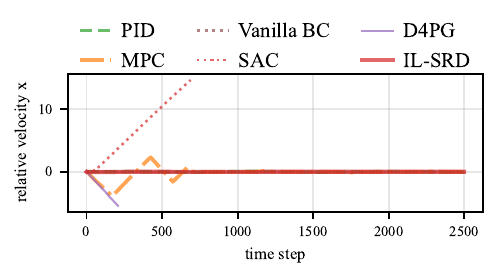}}\hfill
    \subcaptionbox{}{\includegraphics[width=0.33\textwidth]{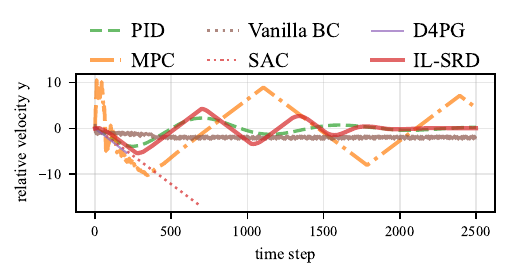}}\hfill
    \subcaptionbox{}{\includegraphics[width=0.33\textwidth]{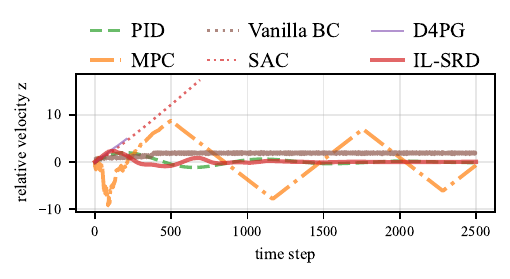}}
    \subcaptionbox{}{\includegraphics[width=0.33\textwidth]{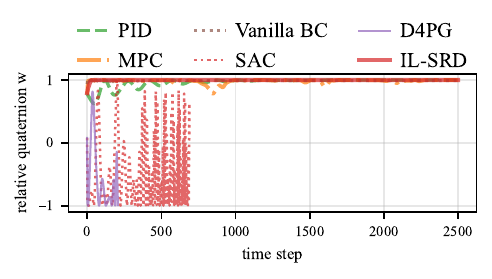}}\hfill
    \subcaptionbox{}{\includegraphics[width=0.33\textwidth]{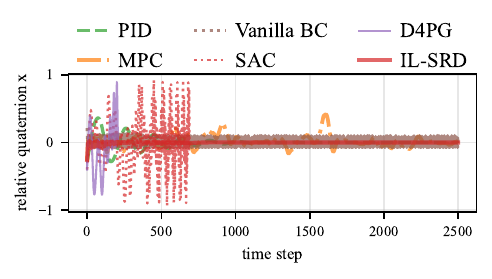}}\hfill
    \subcaptionbox{}{\includegraphics[width=0.33\textwidth]{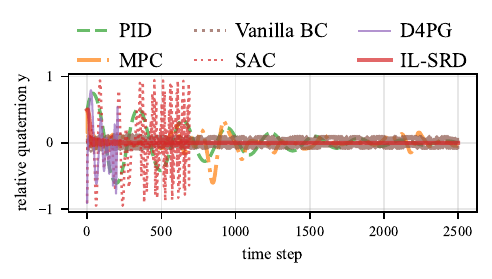}}
    \subcaptionbox{}{\includegraphics[width=0.33\textwidth]{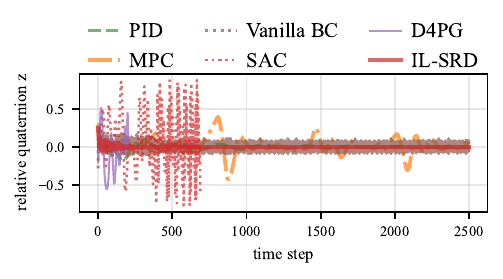}}\hfill
    \subcaptionbox{}{\includegraphics[width=0.33\textwidth]{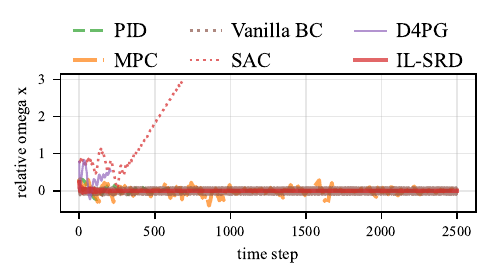}}\hfill
    \subcaptionbox{}{\includegraphics[width=0.33\textwidth]{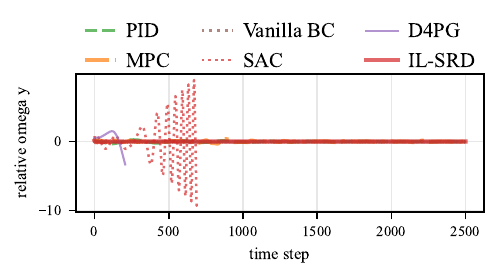}}
    \subcaptionbox{}{\includegraphics[width=0.33\textwidth]{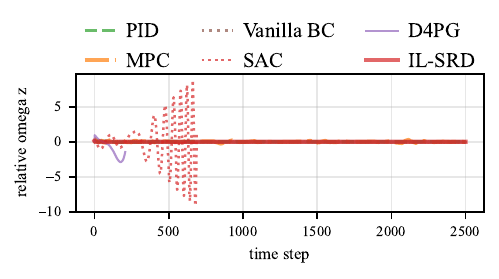}}
    
    \caption{\small Spacecraft rendezvous and docking error diagrams for different control methods under disturbed conditions, (a)--(c) are the relative distance, (d)--(e) are the relative velocities, (g)--(j) are the relative attitude represented by quaternions, (k)--(m) are the relative angular velocities}
    \label{fig:all_results_robust}
\end{figure}

Notably, although the expert demonstrations are generated by an MPC controller, IL-SRD exhibits superior robustness under unseen disturbances. This indicates that the proposed framework successfully distills the underlying control behavior of MPC while discarding its sensitivity to modeling inaccuracies and prediction errors, thereby achieving enhanced robustness in realistic uncertain environments. In addition, we plot the spacecraft rendezvous and docking errors under disturbed conditions for the aforementioned methods in Fig. \ref{fig:all_results_robust}, which also demonstrate the betterment of our method.

\subsection{Ablation Study}
To further demonstrate the effectiveness of the proposed algorithm, we conduct ablation studies to examine how the length of the action sequence and different decoder targets affect the performance of IL-SRD. In addition, an ablation study is conducted to verify the effectiveness of temporal aggregation. All ablation study results are summarized in Table \ref{table4}. The first four rows denote the performance of IL-SRD with different action sequence lengths. Zero Initial Target indicates that the decoder target is fixed to zero, whereas Learnable Target treats the decoder target as a set of learnable parameters. As training proceeds, the target gradually approaches a more suitable parameter configuration.
\begin{table}[!htbp]
\centering
\caption{Experimental results of ablation study}
\resizebox{\textwidth}{!}{%
\begin{tabular}{ccccc}
\hline
Ablation Studies      & SEC               & ATTP               & ATRP              & ESR                \\ \hline
IL-SRD(100)           & 7.855 $\pm$ 0.624 & 38.103 $\pm$ 8.188 & 0.012 $\pm$ 0.008 & -0.478 $\pm$ 0.249 \\
IL-SRD(200)           & 7.294 $\pm$ 0.275 & 26.059 $\pm$ 5.301 & 0.006 $\pm$ 0.001 & -0.396 $\pm$ 0.252 \\
IL-SRD(400)           & 4.448 $\pm$ 0.046 & 4.780 $\pm$ 0.104  & 0.002 $\pm$ 0.001 & -0.301 $\pm$ 0.328 \\
IL-SRD(600)           & 4.060 $\pm$ 0.225 & 1.622 $\pm$ 0.577  & 0.001             & -0.512 $\pm$ 0.472 \\
Zero Initial Target   & 0.820 $\pm$ 0.279 & 3.311 $\pm$ 0.840  & 0.003             & \textbf{-0.117 $\pm$ 0.276} \\
Learnable Target      & 3.812 $\pm$ 0.524 & 2.697 $\pm$ 0.597  & 0.001             & -0.286 $\pm$ 0.385 \\
w/o temporal aggregation & \textbf{0.313 $\pm$ 0.058} & 79.011 $\pm$ 0.592 & 0.008 $\pm$ 0.001 & -1.534 $\pm$ 0.123 \\
IL-SRD(Ours)          & 4.166 $\pm$ 0.269 & \textbf{1.229 $\pm$ 0.185}  & \textbf{0} & -0.266 $\pm$ 0.309 \\ \hline
\end{tabular}%
}
\label{table4}
\end{table}

The length of action sequences significantly affects the performance of the proposed algorithm. Table \ref{table4} shows that smaller sequence lengths fail to achieve convergence. With shorter action sequences, the learning process captures a fragmented state-to-action mapping rather than the global structure of expert trajectories. In addition, the limited temporal context amplifies the impact of unknown sensor errors, leading to oscillatory patterns that result in a less stable control process during evaluation. Excessively long sequence lengths (e.g., 600) also degrade performance. This is because the attention weights are diluted across a larger number of actions. Instead of learning the global features of a complete control cycle, the model tends to learn averaged features spanning multiple cycles.

Among different settings for the decoder target, zero initial target achieves the worst performance. This is because, without the anchored decoder target, the only prior information received by the decoder is a distribution of the future actions of the expert data. Such prior information is insufficient to ensure reliable convergence under limited training episodes. In this case, the future action sequence generated by the policy lacks temporal consistency with the current state, causing the predicted action sequence to be more deviated from the expert data thereby resulting in degraded performance. A similar explanation also applies to the learnable target, however, as learnable target can gradually approach an optimal target. However, with continuously changing training samples, it becomes difficult to converge to a stable optimal target. As a result, a more generic target weakens the temporal dependencies between the target and the predicted action sequence, which negatively affects performance.

We further investigate the effectiveness of the proposed temporal aggregation mechanism. As shown in the results, removing temporal aggregation leads to a pronounced degradation in control performance. This is because the mechanism progressively combines overlapping predicted actions from different action sequences at each timestep, thereby reinforcing temporal consistency across the predicted action sequences. Moreover, unknown sensor errors are effectively smoothed through this aggregation process, since the executed control at each timestep is obtained as a weighted combination of multiple action predictions rather than relying on a single estimate.

\section{Conclusion} \label{sec:conclusion}
In this paper, we propose an imitation learning framework for model-free intelligent spacecraft rendezvous and docking control. An anchored decoder target mechanism is introduced to maintain consistency between the current state and the predicted action sequence, thereby reducing distribution shift and suppressing implausible long-horizon predictions. In addition, a temporal aggregation mechanism is incorporated to further improve control smoothness. Experimental results show that the proposed IL-SRD algorithm achieves performance comparable to expert demonstrations and exhibits strong robustness under unknown disturbances.

However, the proposed algorithm still has certain limitations. The achieved control accuracy remains insufficient for spacecraft rendezvous and docking tasks that require extremely high terminal precision during close-proximity operations. In addition, the current framework relies on state-to-state prediction with low-level servo control, which may introduce additional errors and latency. Our future work will focus on improving terminal control precision and extending the framework to a direct state-to-action formulation, enabling more accurate, efficient, and responsive control for time-critical docking maneuvers.

\section*{Acknowledgement}
This work was kindly supported by Joint Funds of the National Natural Science Foundation of China through grant No. U23A20346 and National Natural Science Foundation of China through grant No. 62403162.


\bibliographystyle{elsarticle-num}
\bibliography{sample}

\end{document}